\documentclass[final]{cvpr}

\usepackage{times}
\usepackage{epsfig}
\usepackage{graphicx}
\usepackage{amsmath}
\usepackage{amssymb}
\usepackage{subcaption}
\usepackage{multirow}

\usepackage[pagebackref,breaklinks,colorlinks]{hyperref}

\def\cvprPaperID{20} 
\def\confYear{CVPR 2021}

\begin{document}

\title{Vision-based Neural Scene Representations for Spacecraft}

\author{Anne Mergy*\\
{\tt\small anne.mergy@esa.int}
\and
Gurvan Lecuyer*\\
{\tt\small gurvan.lecuyer@esa.int}
\and
Dawa Derksen\\
{\tt\small dawa.derksen@esa.int}
\and
Dario Izzo\\
{\tt\small dario.izzo@esa.int} \\
\small European Space Agency Noordwijk, 2201 AZ, The Netherlands\\
}

\maketitle

\begin{abstract}
   In advanced mission concepts with high levels of autonomy, spacecraft need to internally model the pose and shape of nearby orbiting objects.
   Recent works in neural scene representations show promising results for inferring generic three-dimensional scenes from optical images. Neural Radiance Fields (NeRF) have shown success in rendering highly specular surfaces using a large number of images and their pose. More recently, Generative Radiance Fields (GRAF) achieved full volumetric reconstruction of a scene from unposed images only, thanks to the use of an adversarial framework to train a NeRF.
   In this paper, we compare and evaluate the potential of NeRF and GRAF to render novel views and extract the 3D shape of two different spacecraft, the Soil Moisture and Ocean Salinity satellite of ESA's Living Planet Programme and a generic cube sat. Considering the best performances of both models, we observe that NeRF has the ability to render more accurate images regarding the material specularity of the spacecraft and its pose. For its part, GRAF generates precise novel views with accurate details even when parts of the satellites are shadowed while having the significant advantage of not needing any information about the relative pose. 
\end{abstract} 

\section{Introduction}
\label{sec:intro}

The reconstruction of the pose and shape of an orbiting object from a sequence of two-dimensional images is an important part of several mission concepts that have emerged in the last decade. 
In the context of active debris removal missions~\cite{liou2011active, palmerini2016guidelines}, the rendezvous, capture 
(or attachment), and deorbit mission phases can leverage the knowledge of the target debris shape.
During the design of the ESA e-deorbit mission~\cite{biesbroek2017deorbit}, for example, de-orbiting techniques based on clamping mechanisms were considered. This was made possible as the target object, the Envisat satellite, while non-cooperative, is known to a high level of details~\cite{virgili2014investigation}. 
The value of computing, possibly on-board, the shape of a generic, unknown, debris is then immediately understood as to allow autonomous operations of the chosen de-orbit strategy.
A similar scenario is that of generic close proximity operations, such as those encountered during rendezvous and docking or formation flight. 
In the case of the Automated Transfer Vehicle operations, the relative pose of the two docking platforms could be determined thanks to the prior detailed knowledge of a specific 3D feature on-board the target~\cite{casonato2004visual}, while during the Prisma mission~\cite{noteborn2011flight} active visual features were used to facilitate a similar task. The efficient and automated determination of poses is actively studied in similar contexts and in~\cite{d2014pose}, for example, the spacecraft pose is determined to a high degree of precision from a single two-dimensional image assuming a detailed knowledge of the spacecraft model. In~\cite{kisantal2020satellite, black2021real} pose reconstruction was carried out from a large dataset of two dimensional images without explicit use of a model.

Space Situational Awareness~\cite{flohrer2017space}, in-orbit servicing, manufacturing and even recycling are also relevant and active fields of aerospace research where the acquisition of information about a generic target structural integrity, function and pose is pursued. The ESA's OMAR (On-orbit Manufacturing Assembly and Recycling) initiative or NASA's OSAM (On-orbit Servicing, Assembly, and Manufacturing Servicing)  shows the commitment of space Agencies to provide future missions with such a capability.

\begin{figure*}[t]
    \centering
    \includegraphics[width=1.\textwidth]{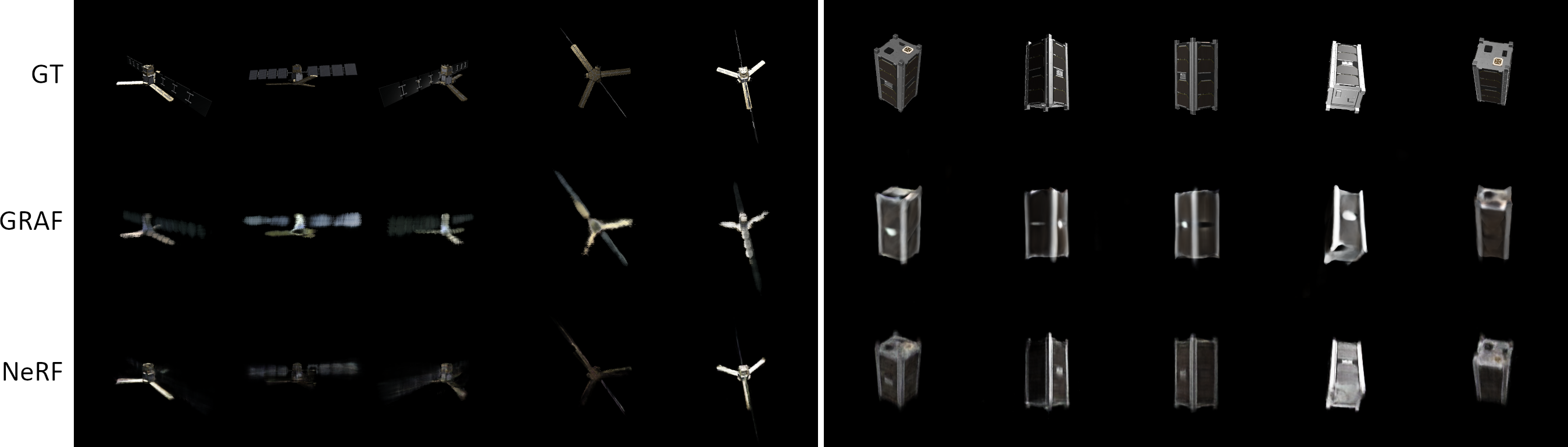}
    \caption{A Neural Radiance Field (NeRF)~\cite{mildenhall2020nerf} and a Generative Radiance Field (GRAF)~\cite{schwarz2020graf} are trained with synthetic datasets of SMOS and CubeSat satellites (resp.), at image resolution $256 \times 256$ pixels. GT: ground truth. We generate images of the satellites from previously unseen viewpoints. Although it is trained with unposed images, GRAF allows for explicit camera control. NeRF generates more accurate images than GRAF regarding the lighting conditions, the view dependent effects and the pose of the object. Both learn surface specularity.}
    \label{fig:comp1}
\end{figure*}

This work considers the use of a single monocular camera  to reconstruct a full colored three-dimensional model of a target orbiting object. The resulting vision-based system is potentially low cost and mainly passive when compared, for example, to a LiDAR based solution~\cite{volpe2017monocular}.
Following recent advances in computer graphics, we leverage Neural Scene Representations to learn a 3D differentiable model of a space object, using images as input. A recent study proposed the Neural Radiance Field (NeRF)~\cite{mildenhall2020nerf}, which computes a continuous mapping from a 3D location and a 2D viewing direction to density and RGB color values. Unlike most previous stereo-photogrammetry approaches, NeRF models are able to account for view-dependent effects such as surface specularity.
More recently, a generative model for radiance fields has performed successful 3D-aware image synthesis and 3D shape estimation of objects from unposed images. Generative Radiance Fields (GRAF)~\cite{schwarz2020graf} offer explicit control on the camera pose, allowing to generate images of the object from unseen view points. They also have the ability to disentangle the shape of the object from its appearance when considering several similar objects.  
To address the scarce availability of good enough images of orbiting objects, synthetic dataset were produced to simulate the necessary orbital scenes and including two different satellites. On these, we compare the ability of NeRF and GRAF to generate images and recreate the accurate shape of the satellites from unseen viewpoints. Using information on the relative position and orientation of the camera, NeRF is able to identify a non-cooperative object, providing a 3D representation of its shape and color values. GRAF, on the other hand, opens a whole new perspective as it performs the same task using unposed images only, without prior knowledge of the viewing direction.

The paper is structured as follows. In Section~\ref{sec:related},
we review related work in novel view synthesis and volumetric representation learning. Section~\ref{sec:awareimage} describes the two methods chosen for the comparison, namely NeRF and GRAF. Section~\ref{sec:exp} details our experimental study, from our dataset to our results. Finally, we discuss the performance of each method and present future work in Section~\ref{sec:final}.

\section{Related work}
\label{sec:related}

\textbf{Novel view synthesis} is the problem of generating an image from an unseen viewpoint given a set of images of a scene. It is an intensively investigated subject by the computer vision community~\cite{tewari2020state}. This task's main challenges are dealing with occlusions and incomplete illumination, particularly when few observations are available.

Traditional approaches for novel view synthesis are image-based rendering methods~\cite{liu2020auto3d, thies2019image, chen2019monocular} which typically use multiple images as input at runtime and warp seen pixels to the targeted view using a blending function to build the new image. 

Recent methods use Deep Learning to extract 3D properties of an object~\cite{liu2020auto3d, thies2019image, chen2019monocular} or a scene~\cite{chen2019monocular}. Liu et al.~\cite{liu2020auto3d} use convolutional neural network (CNN) to describe an object using multiple viewpoints. However such methods do not provide any 3D shape information or lighting control. 

Another method~\cite{thies2019image} proposed to tackle view-dependent effects such as surface specularity by adding an auto-encoder network to learn and predict those artifacts using a depth map as input. Yet, these methods fail at producing valid images from very different viewpoints than the inputs, making them sensitive to cases with sparse observations.

Recent progress has been made using differentiable rendering for learning-based novel view synthesis~\cite{yan2016perspective, kato2018neural} and pose estimation~\cite{pavlakos2018learning}. By calculating the gradients at each point during ray-marching, a neural network can optimize the 3D representation based on 2D projections alone.

\textbf{Explicit scene representation.} A scene can be represented explicitly using discrete elements such as meshes, voxels or point clouds. 
Mesh-based methods are considered for 3D reconstruction~\cite{kato2018neural, wang2018pixel2mesh, liu2019soft} although they are limited in terms of topology. Those methods produce a realistic outcome but they have a high computational cost and they are limited in the number of vertices which can be predicted by a neural network. This hinders their ability to represent fine details, which require a high amount of mesh elements.

Voxel-based novel view synthesis~\cite{yan2016perspective, tulsiani2017multi, henzler2019escaping} involves taking a set of images as input to a network, which predicts a discrete volumetric representation.
However, these models have poor scaling properties with regards to memory usage, because they grow cubically with spatial resolution. 

Point cloud representations use collections of points to model shapes in the 3D space. Their major downside is that they do not contain any topological information. 
They can be integrated into neural networks and are now a common choice for data representation~\cite{lin2018learning, wiles2020synsin}. Point cloud-based approaches have a lower computation cost than the methods previously introduced but they fail to properly render surface information such as the color of a rendered pixel in case of occlusion.

\textbf{Implicit scene representation.} Since NeRF~\cite{mildenhall2020nerf}, implicit volumetric representations have been gaining attention in the computer vision community~\cite{mescheder2019occupancy, yu2020pixelnerf, martin2020nerf, pumarola2020d, park2020deformable}. In this paradigm, the scene representation is learned in a parametrized manner in a neural network.

For each point in 3D space, the neural network can either be trained to model the presence of an object~\cite{mescheder2019occupancy}, or it can be trained to model the local density of light emitting particles, as in NeRF~\cite{mildenhall2020nerf}. The training is self-supervised using multi-view images. Like explicit learning techniques, pixels of input views are projected in a 3D volume along rays cast from the camera. Neural implicit representations model the scene as a continuous function. A major asset of NeRF is its ability to handle view-dependent effects such as surface specularity. The properties of NeRF and the results that it achieved in previous work make it of particular interest for our application. 

\textbf{Generative Adversarial Networks.} Generative Adversarial Networks (GANs)~\cite{goodfellow2014generative, arjovsky2017wasserstein} are another way to render representations of a scene. They have received increased attention in the last years achieving remarkable results in generating realistic high resolution images. Their adversarial learning scheme has led to significant progress in various tasks, such as authentic looking data generation~\cite{radford2015unsupervised, karras2017progressive, karras2020analyzing}, inpainting~\cite{li2017generative, yeh2017semantic}, image-to-image translation~\cite{isola2017image, zhu2017unpaired} and image super-resolution~\cite{ledig2017photo, wang2018esrgan}. Their advantage over classic computer graphics methods is the ability to generate realistic images that appear very different from the ones provided in the training images. This is enabled by the discriminator, which encourages the generator to produce plausible images, according to what it has learned about the training data distribution.
However, most current GAN-based image synthesis methods currently do not provide 3D shape information. They also do not allow for explicit control of camera parameters such as the viewpoint and for fine control over the scene materials or lighting conditions.

Previous work in 3D-aware GAN training, such as PlatonicGAN~\cite{henzler2019escaping} and HoloGAN~\cite{nguyen2019hologan}, are based on voxel grids to characterize either the object or the feature representation, and are therefore constrained by the scaling limitations mentioned earlier. GRAF, on the other hand, uses a NeRF as a 3D implicit representation of the scene and thanks to the discriminator, it has the ability to learn this representation without the information of the pose of the input images. However, Schwarz et al.~\cite{schwarz2020graf} do not directly compare NeRF and GRAF and therefore do not study the posed vs. unposed aspect. Besides the dataset differences, our paper investigates under which sampling conditions generative methods can replace the need for pose information.

\section{3D aware image synthesis}
\label{sec:awareimage}

We compare NeRF and GRAF on their ability to learn a 3D representation of a satellite from 2D supervision using multi-view images. Both methods are trained without any 3D prior on the target which is an interesting aspect for space applications. NeRF needs the information of the position and orientation of the camera in relation to the object. For the same purpose, GRAF does not require any additional input information and has therefore a significant advantage when it comes to identifying non-cooperative spacecraft.
\subsection{Neural Radiance Fields}
The Neural Radiance Field (NeRF), introduced by Mildenhall et al.~\cite{mildenhall2020nerf}, is a neural network that represents a scene as a continuous mapping of 3D locations and 2D viewing directions to an RGB color and a density value. The radiance field is represented using a Multilayer Perceptron (MLP) optimized to map each input 5D coordinate to its corresponding density and emitted color. NeRF parameters are trained using precomputed camera rays extracted from the pixels of the images composing the dataset along with their corresponding view directions.

In the literature, NeRF and its derivatives have been used to perform novel view synthesis using using mainly synthetic data~\cite{mildenhall2020nerf, yu2020pixelnerf, pumarola2020d} but also using real-world data~\cite{mildenhall2020nerf, martin2020nerf,park2020deformable}. Previous studies showed that NeRF is able to learn scene with complex geometry such as a boat rigging~\cite{mildenhall2020nerf}, virtual characters in various poses~\cite{pumarola2020d}, or an object with thin parts~\cite{yu2020pixelnerf} from the ShapeNet reference dataset~\cite{chang2015shapenet}. Another asset of NeRF is its ability to faithfully learn and generate detailed texture. NeRF relies on ray tracing, a rendering technique in computer graphics that simulate light behavior based on geometrical optics theories~\cite{max1995optical}. This is an important feature which allows NeRF to handle view-dependent effects such as specularity on reflective surfaces as shown in Figure~\ref{fig:comp1}.

For the specific case of a close-proximity operation, the spacecraft that attempts the approach is considered as the camera, while the targeted object in space represents the scene to reconstruct. Challenges are in the diffraction-free lighting conditions which cause strong shadow masking on the object surface~\cite{kisantal2020satellite}. This is a major difference with datasets used in previous work which consider scenes with ambient light where shadows have a limited impact, if any. Moreover, spacecraft are typically covered with highly reflective material such as foils used to protect from the Sun's heat.

One natural limitation of the standard NeRF training is the reliance on pose information. In this work we investigate a different training procedure based on generative adversarial learning, that enables volume reconstruction without explicit pose information.

\begin{figure*}[t]
    \centering
    \includegraphics[width=1.\textwidth]{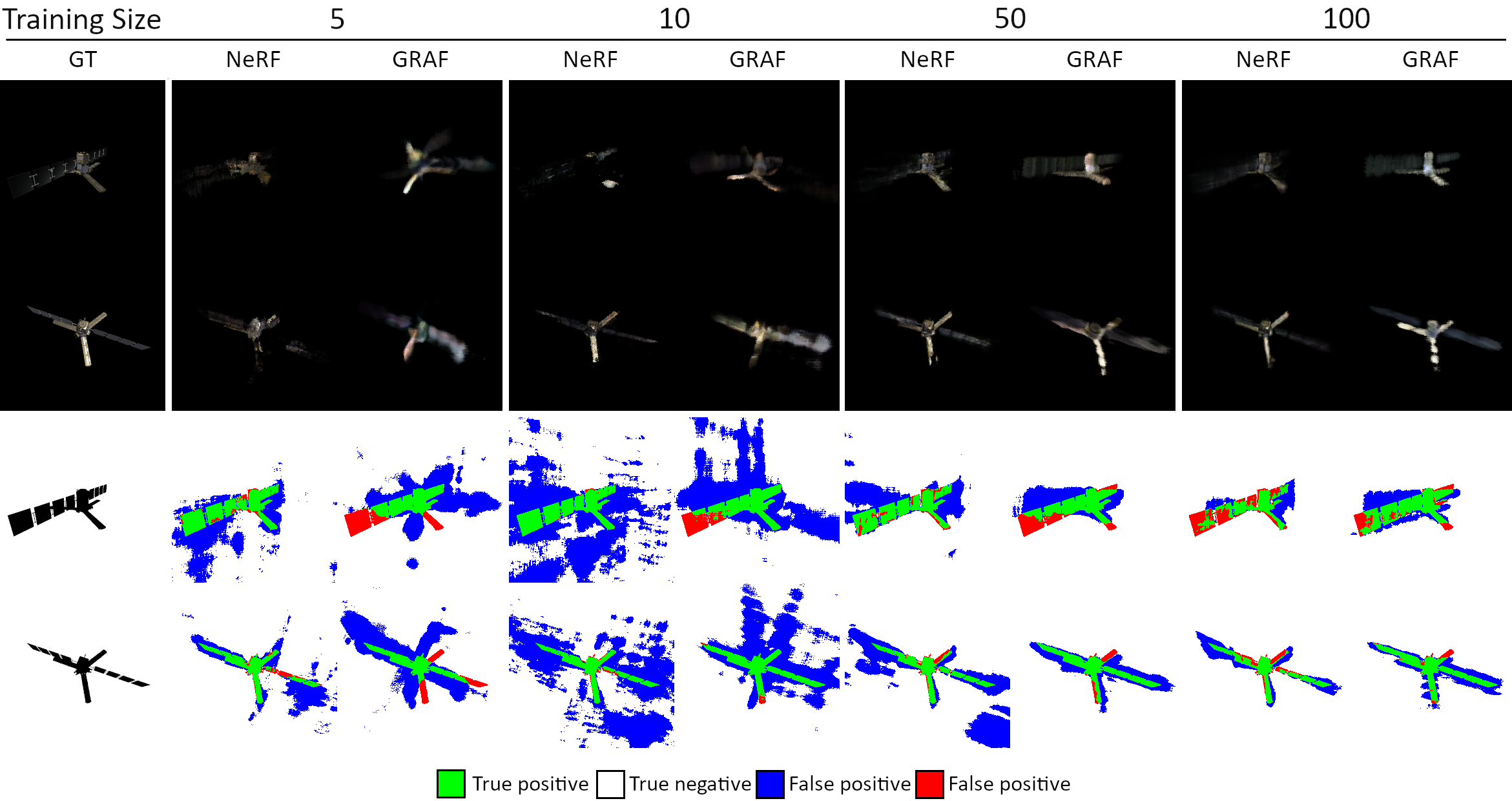}
    \caption{Novel view synthesis and silhouette comparison for NeRF and GRAF trained with training sets of 5, 10, 50 and 100 images of SMOS satellite model. GT stands for ground truth. The top two rows show the reconstructed RGB images for different size of training set for both models. The two last rows show the difference between the true silhouette of the satellite and the presence prediction by the models.}
    \label{fig:comp2}
\end{figure*}

\subsection{Generative Radiance Fields}

Generative Radiance Fields (GRAF)~\cite{schwarz2020graf} use an adversarial training scheme to train a NeRF with a set of unposed images. The internal NeRF representation and the physics-inspired rendering equation are identical to~\cite{mildenhall2020nerf}.

The generator of the GRAF is composed of a radiance field conditioned on several parameters. It takes as input a requested pose that will be used to synthetize an image representation of the scene from that viewpoint during inference. It can optionally be conditioned on shape and appearance codes to enable application to unposed datasets with many different objects. The generator also considers locations of pixels corresponding to a patch. A patch represents a square array of pixels in an image and offers control on the scale of detail that is being learned. The patch position and the distance between the considered pixels in it are used by the generator to synthesize patches of images during the training. Indeed, predicting a color value for every pixel in the target image (in the same batch) is too memory-intensive and restricts application to high-resolution images.

The patch-based discriminator of the GRAF then compares the generated patches to a patch of a real image. It differentiates the real patches from the generated one and this information is used to update the weights of the generator and optimize its loss function. This way, as the training progresses, the generator tries to fool the discriminator by producing pixels matching the probabilistic distribution of the ones located on the real images. Thanks to the discriminator, the NeRF weights are updated to match this distribution by making abstraction of the pose of the images. 
The size of the receptive field of the patch is decreased during the training to start capturing global contours and progressively improve local details. 

During inference, the generator predicts an image of the scene from a specified viewing angle. Changing the shape and appearance codes will correspond to a change in the volume density and color, respectively. While not explored here, this optional feature allows us to consider datasets with a large number of unposed images from slightly different objects. 

\section{Experiments}
\label{sec:exp}
We propose two experiments to assess NeRF and GRAF's capacities to generate novel views and learn a 3D representation of a spacecraft using the synthetic data generated for this study.
In the first experiment, we evaluate the performance of both models using a densely sampled scene of 100 images, in what we would consider ideal conditions. In the second study, we evaluate their performance using sparsely sampled scenes of 5, 10, and 50 images. 
In real space applications, the number of viewpoints for a given target is often limited. The purpose of this experiment is to assess both methods when trained using a sparsely sampled scene.

\subsection{Dataset}
Real space datasets are scarce and often come with limited metadata to supervise the training of learning algorithms. Using synthetic data allows us to produce a large dataset in a controlled environment which eases the production of annotations. We generate the data with the 3D engine Unity using models of two different satellites: a CubeSat and the Soil Moisture and Ocean Salinity (SMOS)~\cite{barre2008smos} satellite.
The generated datasets are publicly available~\cite{mergy_anne_2021_4701174}.
\textit{CubeSat} is a small satellite based on a 3U CubeSat platform. It is a rectangular cuboid shaped of $0.3 \times 0.3 \times 0.9$~m depicted in Figure~\ref{fig:comp1}. Its main structure is made of aluminum and black PCB panels on its sides. For this satellite model, we place the camera at $1$ meter to render the datasets' images. The near and far bounds are fixed at $0.1$ m and $2$ m.

\textit{SMOS} has a more complicated and elongated shape shown in Figure~\ref{fig:comp1}. The main platform has a cubic shape of $0.9 \times 0.9 \times 1.0$~m with solar panels attached on two sides, each $6.5$~m long. The payload is a 3-branch antenna of $3$ meters each placed at $60$ degrees. The structure is covered by golden and silvered foils, which are highly reflective materials. For this satellite model, we place the camera at $10$ meters to render the images. The near and far bounds are fixed at $3$ m and $17$ m due to the solar panel length.

The scene is composed of one satellite, SMOS or CubeSat, with one directional light source fixed with regards to the targeted object. The images are rendered using viewpoints sampled on a full sphere with a unified black background. The images are rendered with a resolution of $1024 \times 1024$ pixels. For each image, the distance to the camera, azimuth and elevation angles are saved as metadata and a depth map is rendered for testing the predicted shape.

We generate training and validation sets containing resp. 5, 10, 50 and 100 images to evaluate the model during training. We also generate a test set of 100 images from different viewing directions than the ones used in the training and validation sets. This common test set will be used to evaluate our models regardless of the number of training images.

\subsection{Metrics}

We use two quantitative metrics to measure the image reconstruction performance: the Peak to Signal Noise Ratio (PSNR) and the Structural Similarity Index (SSIM)~\cite{wang2004image}. The PSNR quantifies the performance of the reconstructed image at pixel level. The SSIM calculates the similarity between two images based on perceived changes such as luminance or contrast.

NeRF and GRAF both learn an implicit representation of the shape of the scene during the training step. Thus it is possible to produce depth maps along with RGB images. We use the Intersection over Union (IoU) metric between the predicted object mask and the reference object mask. This expresses the completeness of the learned shape in a binary detection/non-detection way. To measure the accuracy of the predicted depth, we use the Mean Absolute Error (MAE), expressed in meters, for pixels correctly predicted as the object against their true distances given by our reference 3D model.

\subsection{Model comparison}

NeRF and GRAF are trained using certain aspects of the spatial configuration, in particular the distance between the camera and the object has to be determined along with the near and far bounds defining the volume in which the object is located. NeRF also requires the viewing direction information depicted as elevation and azimuth angles.
The images were downsampled with anti-aliasing to a size of $256~\times~256$ pixels for computational reasons.
For both models, the MLP network is composed of 8 fully-connected layers of 256 channels using ReLU activation functions.

\subsubsection{Comparison after training with 100 images}
\label{sec:exp1}

Figure~\ref{fig:comp1} compares images generated by NeRF and GRAF models trained with 100 training images of SMOS and CubeSat satellites. The images for our qualitative study are chosen to show poses distributed over the camera sphere, highlighting representative view-dependent effects. 
Qualitatively, NeRF generates results similar to the test images. GRAF also achieves good quality results while allowing for explicit camera control, although it is only provided with unposed images without additional information. Symmetries of an object appear to encourage the use of the axis of the natural coordinate system (X, Y, Z) as symmetry planes during the training of the GRAF. Table~\ref{tab:res1} confirms that NeRF leads to a better rendering than GRAF on the generated images as it shows higher PSNR and SSIM values on both SMOS and CubeSat datasets.

\noindent
\begin{table}[h]
\centering
\begin{tabular}{cccccc}
                      & & \multicolumn{4}{c}{SMOS}                                                                  \\ \cline{3-6} 
                      & & PSNR $\uparrow$ & SSIM $\uparrow$ & IoU $\uparrow$ & MAE $\downarrow$ \\ \hline
\multirow{2}{*}{NeRF} & mean & $29.79$ & $0.94$ & $0.94$ & $3.60$ \\
                      & std & $2.15$ & $0.02$ & $0.01$ & $0.23$ \\ \hline
\multirow{2}{*}{GRAF} & mean & $25.61$ & $0.90$ & $0.94$ & $3.81$ \\
                      & std & $3.16$  & $0.02$ & $0.01$ & $0.27$ \\ \hline
\multicolumn{1}{l}{}  & \multicolumn{1}{l}{} & \multicolumn{1}{l}{} & \multicolumn{1}{l}{} & \multicolumn{1}{l}{} & \multicolumn{1}{l}{} \\
                      & & \multicolumn{4}{c}{CubeSat}                                                               \\ \cline{3-6} 
                      & & PSNR $\uparrow$ & SSIM $\uparrow$ & IoU $\uparrow$ & MAE $\downarrow$ \\ \hline
\multirow{2}{*}{NeRF} & mean & $27.62$ & $0.93$ & $0.97$ & $0.53$ \\
                      & std & $2.67$ & $0.01$ & $0.03$ & $0.11$ \\ \hline
\multirow{2}{*}{GRAF} & mean & $19.77$ & $0.84$ & $0.94$ & $0.47$ \\
                      & std & $3.25$ & $0.01$ & $0.01$ & $0.01$ \\ \hline
GRAF -                & mean & $21.05$ & $0.86$ & $0.15$ & $0.52$ \\
overfitting           & std & $3.59$ & $0.02$ & $0.16$ & $0.05$ \\ \hline
\end{tabular}
\caption{Metric values (MAE in meters) comparing RGB and depth predictions with ground truth on the SMOS and CubeSat datasets - NeRF and GRAF are  trained with 100 training images.}
\label{tab:res1}
\end{table}

The scene's simple lighting conditions are learned by both models and rendered correctly on unseen views of the satellite, according to the ground truth. The models properly capture view dependent effects, for instance surface specularity. 
However, as shown on the two last rendered poses of CubeSat in Figure~\ref{fig:comp1}, GRAF can sometimes generate relatively accurate view-dependent effects, as seen on the last image of the line. Sometimes, GRAF generates materials with a blurry aspect. 
Generally speaking, NeRF generates better quality results regarding lighting conditions and view dependent effects. On the other hand and unlike NeRF, GRAF renders an image of the object even when the scene is dark. Even in that case, it can distinguish and learn the shapes of the object. 

The training of the GRAF is monitored qualitatively, by progressively rendering images of the object, and with the computation of the Fréchet Inception Distance (FID)~\cite{heusel2017gans} and the Kernel Inception Distance (KID)~\cite{binkowski2018demystifying} values. On the CubeSat dataset, GRAF renders less accurate images than NeRF. We had to reach a higher number of iterations before it could render the surface specularity as it does in Figure~\ref{fig:comp1}. However, reaching such a high number of iterations means overfitting the dataset. As shown in Table~\ref{tab:res1}, it leads to the generation of a heavily under-regularized depth map which contains many floating artefacts in the background, decreasing the value of the IoU. 
When the training is stopped after a smaller number of iterations, the depth map is more accurate but the model renders a blurry non-detailed RGB image of the CubeSat. A trade-off has to be made between a high number of iterations, leading to a more faithful novel view synthesis, and a small number of iterations leading to the generation of a more accurate depth map. The results of this last trade-off are detailed in the line ``GRAF'' in Table~\ref{tab:res1} while ``GRAF - overfitting'' line shows the metrics of a training with a higher number of iterations. 
On the SMOS dataset, the training of the GRAF was stopped after only 9000 iterations as the model was already capturing details and as we could render accurate images. Both the FID and KID values had almost reached the minimum we could usually observe during training.
On the CubeSat dataset, after 80000 iterations, GRAF renders accurate RGB images but performs poorly compared to NeRF who could almost perfectly learns how to distinguish the spacecraft from the background. If we stop the training after 12000 iterations, the IoU score equals $0.94$ on the same test images and the MAE of the depth prediction is $6$ cm lower than NeRF's. GRAF is also more consistent in its shape estimation as shown by its lower standard deviation.

\begin{figure}[t]
    \centering
    \includegraphics[width=.5\textwidth]{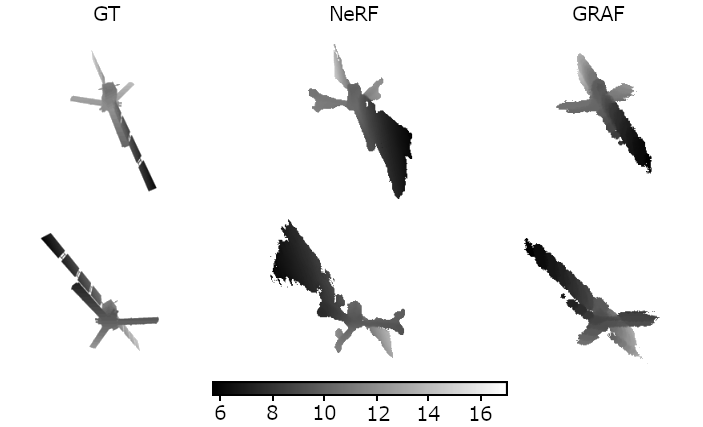}
    \caption{The 3D representation can be extracted as a depth map. NeRF captures fine details of the shape but presents artifacts, while the GRAF rendering is coarse but overall more accurate.}
    \label{fig:comp3}
\end{figure}

On the SMOS dataset, NeRF and GRAF reach a similar performance while learning to distinguish the contours of the spacecraft. As shown in Table~\ref{tab:res1}, they have the same IoU meaning that they have the same completeness for object detection. Figure~\ref{fig:comp3} shows an example of depth estimation for both model on the same viewpoints. Overall, GRAF's depth prediction captures the shape relatively accurately, although the pose is not precise, which shifts it from the ground truth. This explains why the IoU is the same as NeRF although it has a more plausible shape. On the other hand, we notice that in general, NeRF more accurately estimates the distance of the object to the camera explaining why Table~\ref{tab:res1} shows a slightly better mean MAE on the values of the correctly predicted pixels.

\subsubsection{Limiting the amount of training data}
\label{sec:exp2}
\begin{table*}
\centering
\begin{tabular}{llcccccccc}
 \multicolumn{2}{c}{} & \multicolumn{2}{c}{5} & \multicolumn{2}{c}{10} & \multicolumn{2}{c}{50} & \multicolumn{2}{c}{100} \\
 \multicolumn{2}{c}{} & PSNR$\uparrow$ & SSIM$\uparrow$ & PSNR$\uparrow$ & SSIM$\uparrow$ & PSNR$\uparrow$ & SSIM$\uparrow$ & PSNR$\uparrow$ & SSIM$\uparrow$\\ 
\hline
\multirow{2}{*}{NeRF} & mean & 25.81 & 0.90 & 25.46 & 0.90 & 28.35 & 0.93 & 29.79 & 0.94 \\  
 & std & 3.37 & 0.02 & 2.87 & 0.02 & 2.68 & 0.02 & 2.15 & 0.02\\
 \hline
\multirow{2}{*}{GRAF} & mean & 23.26 & 0.85 & 23.64 & 0.86 & 25.58 & 0.89 & 25.61 & 0.90\\ 
 & std & 3.17 & 0.03 & 2.80 & 0.02 & 3.42 & 0.02 & 3.16 & 0.02\\ 
\hline
\hline
 \multicolumn{2}{c}{} & \multicolumn{2}{c}{5} & \multicolumn{2}{c}{10} & \multicolumn{2}{c}{50} & \multicolumn{2}{c}{100} \\
 \multicolumn{2}{c}{} & IoU$\uparrow$ & MAE$\downarrow$ & IoU$\uparrow$ & MAE$\downarrow$ & IoU$\uparrow$ & MAE$\downarrow$ & IoU$\uparrow$ & MAE$\downarrow$\\
\hline
\multirow{2}{*}{NeRF} & mean & 0.88 & 3.75 & 0.72 & 4.54 & 0.86 & 3.71 & 0.94 & 3.60\\ 
  & std & 0.03 & 0.45 & 0.04 & 0.93 & 0.04 & 0.46 & 0.01 & 0.23\\ 
\hline
\multirow{2}{*}{GRAF} & mean & 0.85 & 3.83 & 0.81 & 4.10 & 0.94 & 3.81 & 0.94 & 3.81\\
 & std & 0.03 & 0.48 & 0.04 & 0.45 & 0.01 & 0.29 & 0.01 & 0.27\\
\hline
\end{tabular}
\caption{PSNR (higher is better), SSIM (higher is better), IoU (higher is better), and MAE in meters (lower is better) computed for both NeRF and GRAF trained on 5, 10, 50, and 100 images of the SMOS dataset.}
\label{tab:res2}
\end{table*}
In this second study, we assess the performance of GRAF and NeRF when trained on 5, 10, 50, and 100 images of the SMOS dataset.

Figure~\ref{fig:comp2} compares two poses generated by models trained on respectively 5, 10, 50, and 100 images. The first two rows display the RGB image generated by both models. The last two rows show the difference masks between the true silhouette of the satellite, depicted in the ground truth column, and the predicted presence of the satellite for both models. For these two last rows, green, white, blue and red pixels represent respectively true positives, true negatives, false positives, and false negatives.

The results presented in Section~\ref{sec:exp1} demonstrate that NeRF produces sharper images than GRAF when trained on 100 images. Figure~\ref{fig:comp2} shows that the same is true with smaller training set sizes. The score for PSNR and SSIM shown in Table~\ref{tab:res2} for both models corroborate these observations. As expected, for both GRAF and NeRF, the RGB reconstruction scores increase when trained on an increasing amount of data as shown in Table~\ref{tab:res2}. The use of more data also provides more stable results as the standard deviation of the PSNR and SSIM decrease. 
NeRF is more impacted by sparse views, which cause a difference of almost 4dB for the PSNR and 5\% for the SSIM between the model trained with 100 images and the one trained with 5 images. For GRAF, this difference is only of 1.8 dB for the PSNR and 3\% for the SSIM. Overall, training with fewer input views considerably deteriorates, for both models, the quality of the novel views and the estimated depth.

For models trained using 5 images, the generated views appear blurry, and entire parts of the satellite are not reconstructed. This can be better observed on the difference in silhouettes Figure \ref{fig:comp2}. For both NeRF and GRAF, the results generated when trained on 10 images show fewer missed parts, but the number of false positive increases significantly. Table~\ref{tab:res2} shows that both models reach a higher IoU when trained on 5 data rather than 10, with similar PSNR and SSIM scores. For the training set with 10 images, the image generated by NeRF for the first pose shows a hole in the central structure and in the left solar panel of the satellite. Those artifacts do not appear in the predicted silhouette. For this pose, we believe the model is not able to estimate the color correctly because of the lack of training data. This can also be observed for images generated with NeRF trained with 5 images, and to a lesser extent with 50 and 100 images.

The best results for all of the metrics and for both models are obtained with the models trained on 100 images. However, the results generated by the models trained on 50 images looks very similar in terms of quality. The SSIM score, which is a human perception-based metric, shows similar values when using 50 or 100 training data for both NeRF and GRAF. For GRAF, the PSNR and IoU scores are also very similar when trained on either 50 or 100 data. On the other hand, NeRF scores for these two metrics still improve. By looking carefully at the Figure~\ref{fig:comp2}, we can observe that using 100 training data allows NeRF to find a better estimation for the color and to sharpen the contour of the satellite. This is also shown on the images of differences in silhouette, we can see that the number of false positives decreases when using 100 training data.

The MAE scores in Table~\ref{tab:res2} show that both NeRF and GRAF struggle to capture the real distance between the target and the camera regardless of the training set sizes.

NeRF shows its ability to cope with highly specular effects, even with few images for training. In particular, we observe in Figure~\ref{fig:comp2}, on the second pose, that starting from using 10 images, the specular effects begin to appear on the golden foils on the bottom part of the model. For GRAF, we can observe that these effects appear when trained with 50 images.

The main asset of GRAF compared to NeRF is that it does not need prior knowledge of the pose for training. Figure~\ref{fig:comp2} demonstrates that when trained with 5 images, GRAF is not able to reconstruct coherent images for the required viewpoints. This issue tends to disappear when more data is used for training. However, a slight shift in the pose can still be observed, which penalizes the PSNR and SSIM scores measured for GRAF.

\section{Discussion}
\label{sec:final}
We have compared NeRF and GRAF models to perform 3D aware image synthesis from images. The purpose of this work was to quantify the difference of performance between NeRF and GRAF on a space-related application. Through our experiments, we observed that NeRF outperforms GRAF when assessed with quantitative metrics. When training was stopped earlier to avoid under-regularization effects in the background, it is interesting to notice that GRAF generates depth maps as accurate as NeRF. From a qualitative point of view, both methods have different strengths and weaknesses: NeRF generally produces sharper images, GRAF achieves better results in darker conditions.

The latter observations do not surprise us as NeRF models, unlike GRAF models, are provided also with information on the object pose with respect to the camera. As a consequence, NeRF models have the ability to directly minimize the error between a predicted pixel color and a true value. On the other hand, GRAF learns the object 3D representation by reproducing the probabilistic distribution of the pixels of the images, and optimizes its loss depending on the ability of the discriminator to distinguish real images from generated ones. 
This confirms our expectation that GRAF would be less accurate at reconstructing a specific pose, as its loss function is not optimized based on a difference between a generated value and a true value at a given pose. This can also explain why GRAF succeeds in representing shadowed parts more accurately than NeRF: reproducing a probabilistic distribution allows us to evaluate the plausibility of the images while NeRF would still evaluate them considering shadows as, essentially, black pixels. 

The results on sparse training sets show that the models do not accurately reproduce specular effects or the 3D shape of the spacecraft. Using more images in the training set greatly improved the results for the novel view synthesis but did not enhance the depth estimation. This latter is a requirement for space applications such as close-proximity operations which demand highly accurate measurements of the spacecraft's surrounding environment. For both experiments, the datasets used contained viewpoints evenly distributed on a sphere around the target. It could be interesting to test both models focusing the viewpoints sampling on a single hemisphere of the target. In such a scenario, the video of the spacecraft approaching the target could be used to provide densely sampled viewpoints of a sub-part of the target.

Training NeRF required access to the relative pose of the camera to the targeted scene. In a simulated environment, this information is easily captured, but it is not trivial to get when dealing with real-world data. In theory, this information could be inferred using Machine Learning methods~\cite{peng2019pvnet, gonzalesl6dnet} which usually rely on the detection of keypoints and thus requires extensive annotation for training. However, those methods are ill-suited for highly specular surfaces like foils. As an extension of this work, it could be interesting to train NeRF with noise-corrupted poses to evaluate its sensibility to the accuracy of the viewing angles.

Observing that GRAF reaches a performance similar to NeRF when generating depth maps of an object, although no pose information is provided, is notable. With the appropriate number of iterations and a sufficient number of training images, GRAF reaches an IoU score comparable to NeRF. This observation puts GRAF as a strong choice due to this unposed particularity which is a crucial advantage for an observing satellite. As GRAF defines its own internal coordinate system, and as it tends to use the natural coordinate system (X, Y, Z), it could be used to create a continuous 3D representation of the object when no pose information is available and initialize a shape with internal labels. The trained GRAF could be inverted to predict the pose associated to the different angle views in the same way as iNeRF~\cite{yen2020inerf}. More traditional training could then simply be used to refine the rendering using a NeRF. This last idea could be investigated to perform pose estimation while rendering an accurate 3D representation of a spacecraft. It would be particularly useful for simultaneous localization and mapping (SLAM) during an operational satellite approach situation. 

\section{Conclusion}

We proposed the use of Neural Radiance Fields and Generative Radiance Fields in the context of novel view synthesis of orbiting spacecraft and compared their performances.
We find that both NeRF and GRAF are able to learn 3D differentiable representations of two orbiting spacecraft having a very distinct geometry. 
Overall, NeRF achieves better performance on our datasets, but GRAF is surprisingly competitive, allowing to control the spacecraft pose when rendering new images, while trained with unposed images.
The two datasets produced and used in this work contained scenes of orbiting satellites rather different from the ones traditionally used by the computer vision community. This introduced several small challenges and opportunities, such as dealing with highly specular materials, strong view-dependent effects, a mostly black background and sharp shadows, that both NeRF and GRAF were able to handle.


{\small
\bibliographystyle{ieee_fullname}
\bibliography{egbib}
}

\end{document}